\documentclass[11pt,a4paper]{article}
\usepackage[hyperref]{acl2019}
\usepackage{times}
\usepackage{latexsym}
\usepackage{graphicx}
\usepackage{amsmath}
\usepackage{amssymb}
\usepackage{url}
\usepackage{color}
\usepackage{multirow}
\usepackage{dsfont}
\usepackage{cleveref}

\aclfinalcopy 

\title{Show, Describe and Conclude: \\ On Exploiting the Structure Information of Chest X-Ray Reports}

\author{Baoyu Jing \quad Zeya Wang \quad Eric Xing\\
 Petuum Inc., USA \\  
  {\texttt \{baoyu.jing, zeya.wang, eric.xing\}@petuum.com} \\}

\date{}

\begin{document}
\maketitle
\begin{abstract}
Chest X-Ray (CXR) images are commonly used for clinical screening and diagnosis. Automatically writing reports for these images can considerably lighten the workload of radiologists for summarizing descriptive findings and conclusive impressions. 
The complex structures between and within sections of the reports pose a great challenge to the automatic report generation. 
Specifically, the section \emph{Impression} is a diagnostic summarization over the section \emph{Findings}; and the appearance of \emph{normality} dominates each section over that of \emph{abnormality}. 
Existing studies rarely explore and consider this fundamental structure information. In this work, we propose a novel framework which exploits the structure information between and within report sections for generating CXR imaging reports. 
First, we propose a two-stage strategy that explicitly models the relationship between \emph{Findings} and \emph{Impression}. 
Second, we design a novel co-operative multi-agent system that implicitly captures the imbalanced distribution between \emph{abnormality} and \emph{normality}. 
Experiments on two CXR report datasets show that our method achieves state-of-the-art performance in terms of various evaluation metrics. 
Our results expose that the proposed approach is able to generate high-quality medical reports through integrating the structure information.


\end{abstract}

\section{Introduction}
\begin{figure}[t]
    \centering
    \includegraphics[width=0.5\textwidth]{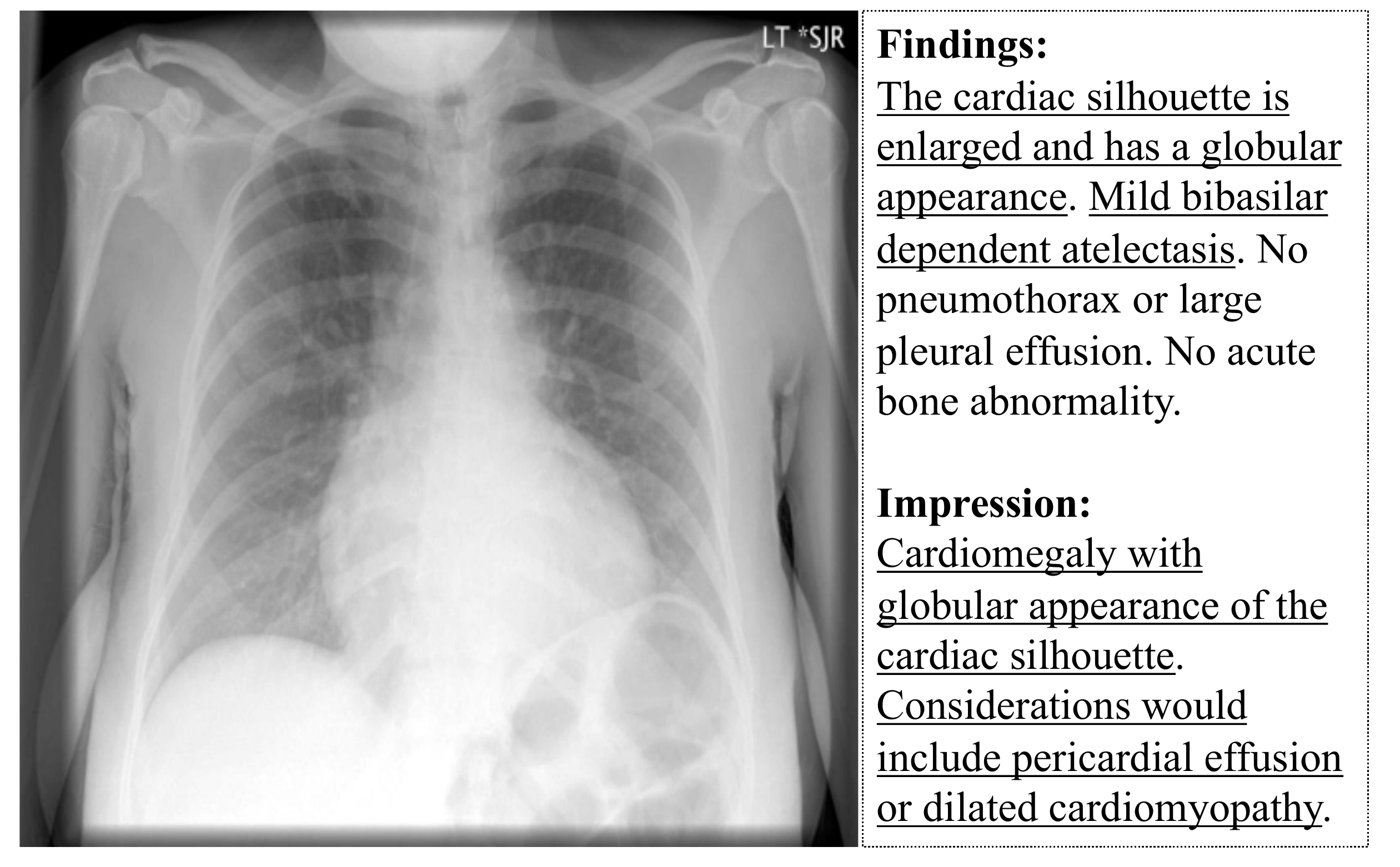}
    \caption{An example of chest X-ray image along with its report. In the report, the \emph{Findings} section records detailed descriptions for normal and abnormal findings; the \emph{Impression} section provides a diagnostic conclusion. The underlined sentence is an abnormal finding.}
    \label{fig:example}
\end{figure}

Chest X-Ray (CXR) image report generation aims to automatically generate detailed findings and diagnoses for given images, which has attracted growing attention in recent years~\cite{wang2018tienet,jing2017automatic,li2018hybrid}. 
This technique can greatly reduce the workload of radiologists for interpreting CXR images and writing corresponding reports. 
In spite of the progress made in this area, it is still challenging for computers to accurately write reports. 
Besides the difficulties in detecting lesions from images, the complex structure of textual reports can prevent the success of automatic report generation. 
As shown in Figure~\ref{fig:example}, the report for a CXR image usually comprises two major sections: \emph{Findings} and \emph{Impression}. 
\emph{Findings} section records detailed descriptions about normal and abnormal findings, such as lesions (e.g. increased lung marking). 
\emph{Impression} section concludes diseases (e.g. pneumonia) from \emph{Findings} and forms a diagnostic conclusion, consisting of abnormal and normal conclusions.



\begin{figure}[h]
    \centering
    \includegraphics[width=0.5\textwidth]{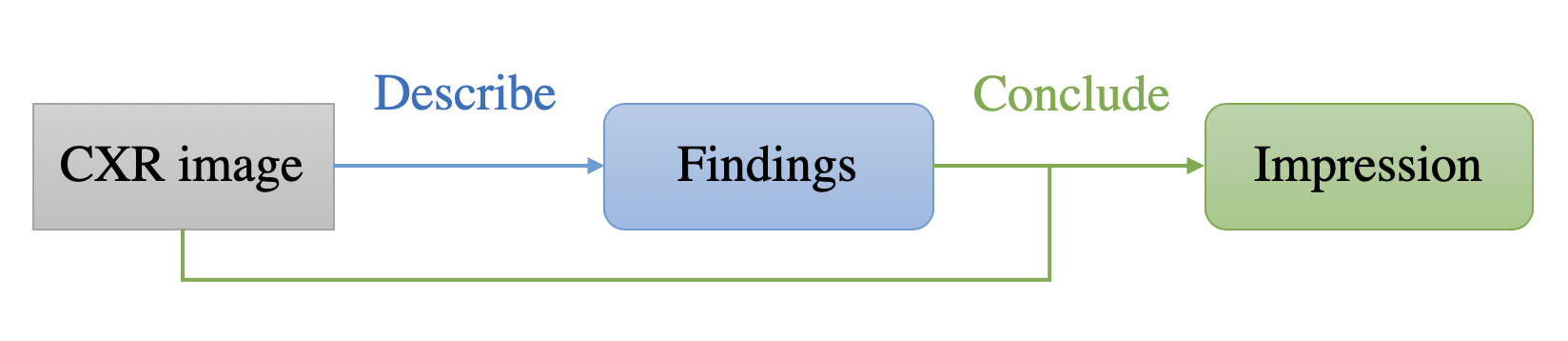}
    \caption{Show, Describe and Conclude.}
    \label{fig:show}
\end{figure}

Existing methods~\cite{wang2018tienet,jing2017automatic,li2018hybrid} ignored the relationship between \emph{Findings} and \emph{Impression}, as well as the different distributions between normal and abnormal findings/conclusions. 
In addressing this problem, we present a novel framework for automatic report generation by exploiting the structure of the reports. 
Firstly, considering the fact that \emph{Impression} is a summarization of \emph{Findings}, we propose a two-stage modeling strategy given in Figure~\ref{fig:show}, where we borrow strength from image captioning task and text summarization task for generating \emph{Impression}.
Secondly, we decompose the generation process of both \emph{Findings} and \emph{Impression} into the following recurrent sub-tasks:
1) examine an area in the image (or a sentence in \emph{Findings}) and decide if an abnormality appears;
2) write detailed (normal or abnormal) descriptions for the examined area.

In order to model the above generation process, we propose a novel Co-operative Multi-Agent System (CMAS), which consists of three agents: Planner (PL), Abnormality Writer (AW) and Normality Writer (NW). 
Given an image, the system will run several loops until PL decides to stop the process.
Within each loop, the agents co-operate with each other in the following fashion:
1) PL examines an area of the input image (or a sentence of \emph{Findings}), and decides whether the examined area contains lesions.
2) Either AW or NW will generate a sentence for the area based on the order given by PL.
To train the system, REINFORCE algorithm~\cite{williams1992simple} is applied to optimize the reward (e.g. BLEU-4 ~\cite{papineni2002bleu}).
To the best of our knowledge, our work is the first effort to investigate the structure of CXR reports.

The major contributions of our work are summarized as follows. 
First, we propose a two-stage framework by exploiting the structure of the reports.
Second, We propose a novel Co-operative Multi-Agent System (CMAS) for modeling the sentence generation process of each section.
Third, we perform extensive quantitative experiments to evaluate the overall quality of the generated reports, as well as the model's ability for detecting medical abnormality terms.
Finally, we perform substantial qualitative experiments to further understand the quality and properties of the generated reports.


\section{Related Work}\label{related_work}
\paragraph{Visual Captioning}
The goal of visual captioning is to generate a textual description for a given image or video. 
For one-sentence caption generation, almost all deep learning methods~\cite{mao2014deep, vinyals2015show, donahue2015long,karpathy2015deep} were based on Convolutional Neural Network (CNN) - Recurrent Neural Network (RNN) architecture.
Inspired by the attention mechanism in human brains, attention-based models, such as visual attention~\cite{xu2015show} and semantic attention~\cite{you2016image}, were proposed for improving the performances.
Some other efforts have been made for building variants of the hierarchical Long-Short-Term-Memory (LSTM) network~\cite{hochreiter1997long} to generate paragraphs~\cite{krause2017hierarchical, yu2016video, liang2017recurrent}. 
Recently, deep reinforcement learning has attracted growing attention in the field of visual captioning~\cite{ren2017deep,rennie2017self,liu2017improved,wang2018video}.
Additionally, other tasks related to visual captioning, (e.g., dense captioning~\cite{johnson2016densecap}, multi-task learning~\cite{pasunuru2017multi}) also attracted a lot of research attention.

\paragraph{Chest X-ray Image Report Generation}
\newcite{shin2016learning} first proposed a variant of CNN-RNN framework to predict tags (location and severity) of chest X-ray images. 
\newcite{wang2018tienet} proposed a joint framework for generating reference reports and performing disease classification at the same time. 
However, this method was based on a single-sentence generation model~\cite{xu2015show}, and obtained low BLEU scores. 
\newcite{jing2017automatic} proposed a hierarchical language model equipped with co-attention to better model the paragraphs, but it tended to produce normal findings. 
Despite \newcite{li2018hybrid} enhanced language diversity and model's ability in detecting abnormalities through a hybrid of template retrieval module and text generation module, manually designing templates is costly and they ignored the template's change over time.

\paragraph{Multi-Agent Reinforcement Learning}
The target of multi-agent reinforcement learning is to solve complex problems by integrating multiple agents that focus on different sub-tasks.
In general, there are two types of multi-agent systems: independent and cooperative systems~\cite{tan1993multi}. 
Powered by the development of deep learning, deep multi-agent reinforcement learning has gained increasing popularity.
\newcite{tampuu2017multiagent} extended Deep Q-Network (DQN)~\cite{mnih2013playing} into a multi-agent DQN for Pong game; \newcite{foerster2016learning, sukhbaatar2016learning} explored communication protocol among agents; \newcite{zhang2018fully} further studied fully decentralized multi-agent system. 
Despite these many attempts, the multi-agent system for long paragraph generation still remains unexplored.

\begin{figure*}
    \centering
    \includegraphics[width=\textwidth]{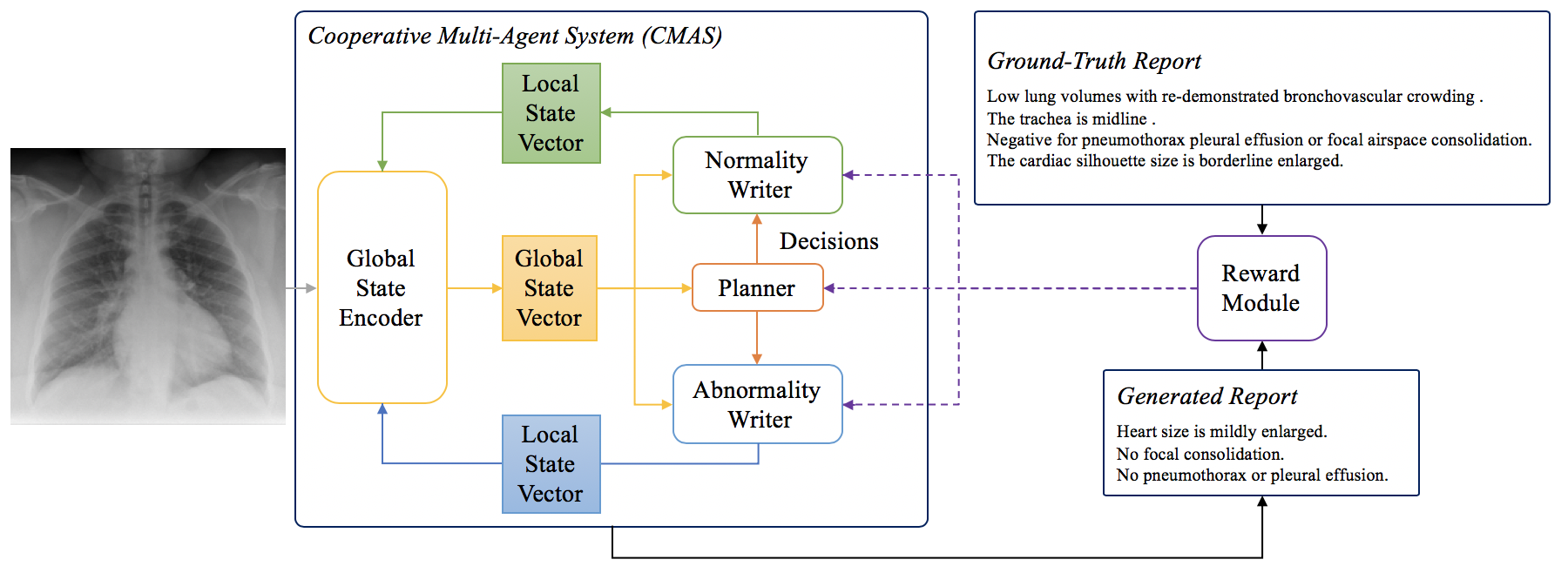}
    \caption{Overview of the proposed Cooperative Multi-Agent System (CMAS).}
    \label{fig:overview}
\end{figure*}

\section{Overall Framework}\label{framework}
As shown in Figure~\ref{fig:show}, the proposed framework is comprised of two modules: Findings and Impression. 
Given a CXR image, the Findings module will examine different areas of the image and generate descriptions for them.
When findings are generated, the Impression module will give a conclusion based on findings and the input CXR image.
The proposed two-stage framework explicitly models the fact that \emph{Impression} is a conclusive summarization of \emph{Findings}.

Within each module, we propose a Co-operative Multi-Agent System (CMAS) (see Section ~\ref{approach}) to model the text generation process for each section. 

\section{Co-operative Multi-Agent System}\label{approach}
\subsection{Overview}
The proposed Co-operative Multi-Agent System (CMAS) consists of three agents: Planner (PL), Normality Writer (NW) and Abnormality Writer (AW). 
These agents work cooperatively to generate findings or impressions for given chest X-ray images. 
PL is responsible for determining whether an examined area contains abnormality, while NW and AW are responsible for describing normality or abnormality in detail (Figure~\ref{fig:overview}).

The generation process consists of several loops, and each loop contains a sequence of actions taken by the agents. 
In the $n$-th loop, the writers first share their local states $LS_{n-1,T} = \{w_{n-1, t}\}_{t=1}^{T}$ (actions taken in the previous loop) to form a shared global state $GS_{n}=(I, \{s_i\}_{i=1}^{n-1})$, where $I$ is the input image, $s_i$ is the $i$-th generated sentence, and $w_{i,t}$ is the $t$-th word in the $i$-th sentence of length $T$.  
Based on the global state $GS_{n}$, PL decides whether to stop the generation process or to choose a writer (NW or AW) to produce the next sentence $s_n$ . 
If a writer is selected, then it will refresh its memory by $GS_{n}$ and generate a sequence of words $\{w_{n,t}\}_{t=1}^T$ based on the sequence of local state $LS_{n, t}=\{w_{n, 1},\cdots, w_{n, t-1}$\}.

Once the generation process is terminated, the reward module will compute a reward by comparing the generated report with the ground-truth report. 
Given the reward, the whole system is trained via REINFORCE algorithm~\cite{williams1992simple}.

\subsection{Policy Network}
\subsubsection{Global State Encoder}\label{GSE}
During the generation process, each agent will make decisions based on the global state $GS_{n}$. 
Since $GS_{n}$ contains a list of sentences $\{s_i\}_{i=1}^{n-1}$, a common practice is to build a hierarchical LSTM as Global State Encoder (GSE) for encoding it. 
Equipping such an encoder with an excessive number of parameters for each agent in CMAS would be computation-consuming. 
We address this problem in two steps. 
First, we tie weights of GSE across the three agents. 
Second, instead of encoding previous sentences from scratch, GSE dynamically encodes $GS_{n}$ based on $GS_{n-1}$.
Specifically, we propose a single layer LSTM with soft-attention~\cite{xu2015show} as GSE. 
It takes a multi-modal context vector $\mathbf{ctx}_n\in\mathbb{R}^H$ as input, which is obtained by jointly embedding sentence $s_{n-1}$ and image $I$ to a hidden space of dimension $H$, and then generates the global hidden state vector $\mathbf{gs}_n\in\mathbb{R}^H$ for the $n$-th loop by:
\begin{equation}
    \mathbf{gs}_n = \text{LSTM}(\mathbf{gs}_{n-1}, \mathbf{ctx}_n)
\end{equation}

We adopt a visual attention module for producing context vector $\mathbf{ctx}_n$, given its capability of capturing the correlation between languages and images~\cite{lu2017knowing,xu2015show}. 
The inputs to the attention module are visual feature vectors $\{\mathbf{v}_p\}_{p=1}^{P}\in\mathbb{R}^C$ and local state vector $\mathbf{ls}_{n-1}$ of sentence $s_{n-1}$.
Here, $\{\mathbf{v}_p\}_{p=1}^{P}$ are extracted from an intermediate layer of a CNN, $C$ and $p$ are the number of channels and the position index of $\mathbf{v}_p$.
$\mathbf{ls}_{n-1}$ is the final hidden state of a writer (defined in section~\ref{writer}). 
Formally, the context vector $\mathbf{ctx}_n$ is computed by the following equations:
\begin{equation}\label{va1}
    \mathbf{h}_{p} = \tanh(\mathbf{W}_h[\mathbf{ls}_{n-1}; \mathbf{gs}_{n-1}])
\end{equation}
\begin{equation}\label{va2}
    \alpha_{p} = \frac{\exp(\mathbf{W}_{att}\mathbf{h}_{p})}{\sum_{q=1}^{P}\exp(\mathbf{W}_{att}\mathbf{h}_{q})}
\end{equation}
\begin{equation}\label{va3}
    \mathbf{v}_{att} = \sum_{p=1}^P\alpha_{p}\mathbf{v}_p
\end{equation}
\begin{equation}\label{va4}
    \mathbf{ctx}_{n} = \tanh(\mathbf{W}_{ctx}[\mathbf{v}_{att}; \mathbf{ls}_{n-1}])
\end{equation}
where $\mathbf{W}_{h}$, $\mathbf{W}_{att}$ and $\mathbf{W}_{ctx}$ are parameter matrices; $\{\alpha_p\}_{p=1}^P$ are weights for visual features; and $[;]$ denotes concatenation operation.

At the beginning of the generation process, the global state is $GS_1=(I)$.
Let $\mathbf{\bar{v}}=\frac{1}{P}\sum_{i=1}^{P}\mathbf{v}_i$, the initial global state $\mathbf{gs}_0$ and cell state $\mathbf{c}_{0}$ are computed by two single-layer neural networks:
\begin{align}
    \mathbf{gs}_0 &= \tanh(\mathbf{W}_{gs}\mathbf{\bar{v}})\\
    \mathbf{c}_{0} &= \tanh(\mathbf{W}_{c}\mathbf{\bar{v}})
\end{align}
where $\mathbf{W}_{gs}$ and $\mathbf{W}_{c}$ are parameter matrices.

\subsubsection{Planner}
After examining an area, Planner (PL) determines: 1) whether to terminate the generation process; 2) which writer should generate the next sentence.
Specifically, besides the shared Global State Encoder (GSE), the rest part of PL is modeled by a two-layer feed-forward network:
\begin{align}
    \mathbf{h}_n &= \tanh(\mathbf{W}_2\tanh(\mathbf{W}_1\mathbf{gs}_n))\\
    idx_n &= \arg \text{max}(\text{softmax}(\mathbf{W}_3\mathbf{h}_n))
\end{align}
where $W_1$, $W_2$, and $W_3$ are parameter matrices; $idx_n\in\{0, 1, 2\}$ denotes the indicator, where $0$ is for STOP, $1$ for NW and $2$ for AW. Namely, if $idx_n=0$, the system will be terminated; else, NW ($idx_n=1$) or AW ($idx_n=2$) will generate the next sentence $s_n$.

\subsubsection{Writers}\label{writer}
The number of normal sentences is usually 4-12 times to the number of abnormal sentences for each report. 
With this highly unbalanced distribution, using only one decoder to model all of the sentences would make the generation of normal sentences dominant.
To solve this problem, we design two writers, i.e., Normality Writer (NW) and Abnormality Writer (AW), to model normal and abnormal sentences. 
Practically, the architectures of NW and AW can be different. In our practice, we adopt a single-layer LSTM for both NW and AW given the principle of parsimony. 

Given a global state vector $\mathbf{gs}_n$, CMAS first chooses a writer for generating a sentence based on $idx_n$.
The chosen writer will re-initialize its memory by taking $\mathbf{gs}_n$ and a special token BOS (Begin of Sentence) as its first two inputs. 
The procedure for generating words is:
\begin{align}
    \mathbf{h}_{t} &= \text{LSTM}(\mathbf{h}_{t-1}, \mathbf{W_e}\mathbf{y}_{w_{t-1}})\\
    \mathbf{p}_t &= \text{softmax}(\mathbf{W}_{out}\mathbf{h}_t)\\
    w_{t} &=\arg\text{max} (\mathbf{p}_t)
\end{align}
where $\mathbf{y}_{w_{t-1}}$ is the one-hot encoding vector of word $w_{t-1}$; $\mathbf{h}_{t-1}, \mathbf{h}_t\in\mathbb{R}^H$ are hidden states of LSTM; $\mathbf{W_e}$ is the word embedding matrix and $\mathbf{W}_{out}$ is a parameter matrix. $\mathbf{p}_t$ gives the output probability score over the vocabulary.

Upon the completion of the procedure (either token EOS (End of Sentence) is produced or the maximum time step $T$ is reached), the last hidden state of LSTM will be used as local state vector $\mathbf{ls}_n$, which will be fed into GSE for generating next global state vector $GS_{n+1}$.

\subsection{Reward Module}
We use BLEU-4~\cite{papineni2002bleu} to design rewards for all agents in CMAS. 
A generated paragraph is a collection $(\mathbf{s}^{ab}, \mathbf{s}^{nr})$ of normal sentences $\mathbf{s}^{nr}= \{s^{nr}_1, \dots, s^{nr}_{N_{nr}}\}$ and abnormal sentences $\mathbf{s}^{ab} = \{s^{ab}_1, \dots, s^{ab}_{N_{ab}}\}$, where $N_{ab}$ and $N_{nr}$ are the number of abnormal sentences and the number of normal sentences, respectively.
Similarly, the ground truth paragraph corresponding to the generated paragraph $(\mathbf{s}^{ab}, \mathbf{s}^{nr})$ is $(\mathbf{s}^{\ast ab}, \mathbf{s}^{\ast nr})$.

We compute BLEU-4 scores separately for abnormal and normal sentences.
For the first $n$ generated abnormal and normal sentences, we have:
\begin{align}
    f(s^{ab}_n) &= \text{BLEU}(\{s^{ab}_1,\cdots, s^{ab}_n\}, \mathbf{s}^{\ast ab})\\
    f(s^{nr}_n) &= \text{BLEU}(\{s^{nr}_1,\cdots, s^{nr}_n\}, \mathbf{s}^{\ast nr})
\end{align}

Then, the immediate reward for $s_n$ ($s^{ab}_n$ or $s^{nr}_n$) is $r(s_n) = f(s_n) - f(s_{n-1})$.
Finally, the discounted reward for $s_n$ is defined as:
\begin{equation}\label{eq_reward}
    R(s_n) = \sum_{i=0}^{\infty}\gamma^ir(s_{n+i})
\end{equation}
where $\gamma\in[0, 1]$ denotes discounted factor, and $r(s_1) = \text{BLEU}(\{s_1\}, \mathbf{s}^{\ast})$.


\subsection{Learning} \label{Learning}
\subsubsection{Reinforcement Learning}
Given an input image $I$, three agents (PL, NW and AW) in CMAS work simultaneously to generate a paragraph $\mathbf{s}$ = $\{s_1, s_2, \dots, s_N\}$ with the joint goal of maximizing the discounted reward $R(s_n)$ (Equation~\ref{eq_reward}) for each sentence $s_n$.

The loss of a paragraph $\mathbf{s}$ is negative expected reward:
\begin{equation}\label{eq_loss_all}
    L(\theta) = -\mathbb{E}_{n,s_n\sim\pi_\theta}[R(s_n)]
\end{equation}
where $\pi_\theta$ denotes the entire policy network of CMAS. 
Following the standard REINFORCE algorithm~\cite{williams1992simple}, the gradient for the expectation $\mathbb{E}_{n, s_n\sim\pi_\theta}[R(s_n)]$ in Equation~\ref{eq_loss_all} can be written as:
\begin{equation}\label{eq_grad}
    \nabla_\theta L(\theta) = \mathbb{E}_{n, s_n\sim\pi_\theta}[R(s_n)\nabla_\theta -\log \pi_\theta(s_n, idx_n)]
\end{equation}
where $-\log\pi_{\theta}(s_n, idx_n)$ is joint negative log-likelihood of sentence $s_n$ and its indicator $idx_n$, and it can be decomposed as:
\begin{equation}
\begin{split}
    &-\log\pi_\theta(s_n, idx_n) \\
    =& \mathds{1}_{\{idx_n=AW\}}L_{AW} + \mathds{1}_{\{idx_n=NW\}}L_{NW} + L_{PL} \\
    =& - \mathds{1}_{\{idx_n=AW\}}\sum_{t=1}^{T}\log p_{AW}(w_{n,t}) \\ 
    &- \mathds{1}_{\{idx_n=NW\}}\sum_{t=1}^{T}\log p_{NW}(w_{n,t})  \\
    &-\log p_{PL}(idx_n)
\end{split}
\end{equation}
where $L_{AW}$, $L_{NW}$ and $L_{PL}$ are negative log-likelihoods; $p_{AW}$, $p_{NW}$ and $p_{PL}$ are probabilities of taking an action; $\mathds{1}$ denotes indicator function.

Therefore, Equation~\ref{eq_grad} can be re-written as:
\begin{equation}\label{loss_rl}
\begin{split}
    \nabla_\theta L(\theta) &=
    \mathbb{E}_{n, s_n\sim\pi_\theta}[R(s_n)(\mathds{1}_{\{idx_n=AW\}}\nabla L_{AW}\\
    &+ \mathds{1}_{\{idx_n=NW\}}\nabla L_{NW} + \nabla L_{PL})]
\end{split}
\end{equation}

\subsubsection{Imitation Learning}
It is very hard to train agents using reinforcement learning from scratch, therefore a good initialization for policy network is usually required~\cite{bahdanau2016actor, silver2016mastering, wang2018video}.
We apply imitation learning with cross-entropy loss to pre-train the policy network. 
Formally, the cross-entropy loss is defined as:
\begin{equation}\label{loss_ce}
\begin{split}
    &L_{CE}(\theta) = -\lambda_{PL}\sum_{n=1}^{N}\{\log p_{PL}(idx_n^\ast)\}\\
    -&\lambda_{NW}\sum_{n=1}^{N}\{\mathds{1}_{\{idx_n^\ast=NW\}}\sum_{t=1}^{T}\log p_{NW}(w_{n,t}^\ast)\} \\ 
    -& \lambda_{AW}\sum_{n=1}^{N}\{\mathds{1}_{\{idx_n^\ast=AW\}}\sum_{t=1}^{T}\log p_{AW}(w_{n,t}^\ast)\} \\
\end{split}
\end{equation}
where $w^\ast$ and $idx^\ast$ denote ground-truth word and indicator respectively; $\lambda_{PL}$, $\lambda_{NW}$ and $\lambda_{AW}$ are balancing coefficients among agents; $N$ and $T$ are the number of sentences and the number of words within a sentence, respectively.

\subsection{CMAS for Impression}
Different from the Findings module, the inputs of the Impression module not only contain images $I$ but also the generated findings $\mathbf{f}=\{f_1,f_2,\dots,f_{N_{f}}\}$, where $N_f$ is the total number of sentences. 
Thus, for the Impression module, the $n$-th global state becomes $GS_n=(I, \mathbf{f}, \{s_i\}_{i=1}^{n-1})$. 
The rest part of CMAS for the Impression module is exactly the same as CMAS for the Findings module.
To encode $\mathbf{f}$, we extend the definition of multi-modal context vector $\mathbf{ctx}_n$ (Equation~\ref{va4}) to:
\begin{equation}\label{va24}
    \mathbf{ctx}_{n} = \tanh(\mathbf{W}_{ctx}[\mathbf{v}_{att}; \mathbf{f}_{att}; \mathbf{ls}_{n-1}])
\end{equation}
where $\mathbf{f}_{att}$ is the soft attention~\cite{bahdanau2014neural,xu2015show} vector, which is obtained similar as $\mathbf{v}_{att}$ (Equation~\ref{va2} and~\ref{va3}). 

\section{Experiments}\label{experiments}
\subsection{Datasets}
\paragraph{IU-Xray} Indiana University Chest X-Ray Collection~\cite{demner2015preparing} is a public dataset containing 3,955 fully de-identified radiology reports collected from the Indiana Network for Patient Care, each of which is associated with a frontal and/or lateral chest X-ray images, and there are 7,470 chest X-ray images in total. 
Each report is comprised of several sections: \emph{Impression}, \emph{Findings} and \emph{Indication} etc. We preprocess the reports by tokenizing, converting tokens into lower-cases and removing non-alpha tokens. 


\paragraph{CX-CHR} CX-CHR~\cite{li2018hybrid} is a proprietary internal dataset, which is a Chinese chest X-ray report dataset collected from a professional medical examination institution. 
This dataset contains examination records for 35,500 unique patients, each of which consists of one or multiple chest X-ray images as well as a textual report written by professional radiologists. 
Each textual report has sections such as \emph{Complain}, \emph{Findings} and \emph{Impression}.
The textual reports are preprocessed through tokenizing by ``jieba''\footnote{https://github.com/fxsjy/jieba.}, a Chinese text segmentation tool, and filtering rare tokens. 

For both datasets, we used the same data splits as \citeauthor{li2018hybrid}.

\subsection{Experimental Setup}
\paragraph{Abnormality Term Extraction}
Human experts helped manually design patterns for most frequent medical abnormality terms in the datasets.
These patterns are used for labeling abnormality and normality of sentences, and also for evaluating models' ability to detect abnormality terms.
The abnormality terms in \emph{Findings} and \emph{Impression} are different to some degree.
This is because many abnormality terms in \emph{Findings} are descriptions rather than specific disease names. 
For examples, ``low lung volumes'' and ``thoracic degenerative'' usually appear in \emph{Findings} but not in \emph{Impression}.


\paragraph{Evaluation Metrics} 
We evaluate our proposed method and baseline methods on: BLEU~\cite{papineni2002bleu}, ROUGE~\cite{lin2004rouge} and CIDEr~\cite{vedantam2015cider}. 
The results based on these metrics are obtained by the standard image captioning evaluation tool\footnote{https://github.com/tylin/coco-caption}. 
We also calculate precision and average False Positive Rate (FPR) for abnormality detection in generated textual reports on both datasets.

\paragraph{Implementation Details}
The dimensions of all hidden states in Abnormality Writer, Normality Writer, Planner and shared Global State Encoder are set to 512. 
The dimension of word embedding is also set as 512.

We adopt ResNet-50~\cite{he2016deep} as image encoder, and visual features are extracted from its last convolutional layer, which yields a $7\times7\times2048$ feature map.
The image encoder is pretrained on ImageNet~\cite{deng2009imagenet}).
For the IU-Xray dataset, the image encoder is fine-tuned on ChestX-ray14 dataset~\cite{wang2017chestx}, since the IU-Xray dataset is too small.
For the CX-CHR dataset, the image encoder is fine-tuned on its training set.
The weights of the image encoder are then fixed for the rest of the training process.

During the imitation learning stage, the cross-entropy loss (Equation~\ref{loss_ce}) is adopted for all of the agents, where $\lambda_{PL}$, $\lambda_{AW}$ and $\lambda_{NW}$ are set as 1.0.
We use Adam optimizer~\cite{kingma2014adam} with a learning rate of $5\times10^{-4}$ for both datasets. 
During the reinforcement learning stage, the gradients of weights are calculated based on Equation~\ref{loss_rl}. 
We also adopt Adam optimizer for both datasets and the learning rate is fixed as $10^{-6}$.




\paragraph{Comparison Methods}
For \emph{Findings} section, we compare our proposed method with state-of-the-art methods for CXR imaging report generation:
CoAtt~\cite{jing2017automatic} and HGRG-Agent~\cite{li2018hybrid}, as well as several state-of-the-art image captioning models: CNN-RNN~\cite{vinyals2015show}, LRCN~\cite{donahue2015long}, AdaAtt~\cite{lu2017knowing}, Att2in~\cite{rennie2017self}.
In addition, we implement several ablated versions of the proposed CMAS to evaluate different components in it:
$\text{CMAS}_\text{W}$ is a single agent system containing only one writer, but it is trained on both normal and abnormal findings.
$\text{CMAS}_\text{NW,AW}$ is a simple concatenation of two single agent systems $\text{CMAS}_\text{NW}$ and $\text{CMAS}_\text{AW}$, which are respectively trained on only normal findings and only abnormal findings. 
Finally, we show CMAS's performances with imitation learning (CMAS-IL) and reinforcement learning (CMAS-RL).

For \emph{Impression} section, we compare our method with \newcite{xu2015show}: SoftAtt$_\text{vision}$ and SoftAtt$_\text{text}$, which are trained with visual input only (no findings) and textual input only (no images).
We also report CMAS trained only on visual and textual input: $\text{CMAS}_\text{text}$ and $\text{CMAS}_\text{vision}$.
Finally, we also compare CMAS-IL with CMAS-RL.

\begin{table*}[t!]
\centering
\small
\begin{tabular}{c|l|c c c c c c}
\hline
Dataset & Methods & BLEU-1 & BLEU-2 & BLEU-3 & BLEU-4 & ROUGE & CIDEr\\
\hline
\multirow{10}{*}{CX-CHR}
 & CNN-RNN~\cite{vinyals2015show}   & 0.590 & 0.506 & 0.450 & 0.411 & 0.577 & 1.580\\
 & LRCN~\cite{donahue2015long}      & 0.593 & 0.508 & 0.452 & 0.413 & 0.577 & 1.588\\
 & AdaAtt~\cite{lu2017knowing}      & 0.588 & 0.503 & 0.446 & 0.409 & 0.575 & 1.568\\
 & Att2in~\cite{rennie2017self}     & 0.587 & 0.503 & 0.446 & 0.408 & 0.576 & 1.566\\
 & CoAtt~\cite{jing2017automatic}   & 0.651 & 0.568 & 0.521 & 0.469 & 0.602 & 2.532\\
 & HGRG-Agent~\cite{li2018hybrid}   & 0.673 & 0.587 & 0.530 & 0.486 & 0.612 & 2.895\\
 \cline{2-8}
 & CMAS$_\text{W}$     & 0.659 & 0.585 & 0.534 & 0.497 & 0.627 & 2.564\\
 & CMAS$_\text{NW,AW}$  & 0.657 & 0.579 & 0.522 & 0.479 & 0.585 & 1.532\\
 & CMAS-IL       & 0.663 & 0.592 & 0.543 & 0.507 & 0.628 & 2.475\\
 & CMAS-RL       & \textbf{0.693} & \textbf{0.626} & \textbf{0.580} & \textbf{0.545} & \textbf{0.661} & \textbf{2.900}\\
\hline
\hline
\multirow{10}{*}{IU-Xray}
 & CNN-RNN \cite{vinyals2015show}   & 0.216 & 0.124 & 0.087 & 0.066 & 0.306 & 0.294\\
 & LRCN \cite{donahue2015long}      & 0.223 & 0.128 & 0.089 & 0.067 & 0.305 & 0.284\\
 & AdaAtt~\cite{lu2017knowing}      & 0.220 & 0.127 & 0.089 & 0.068 & 0.308 & 0.295\\
 & Att2in \cite{rennie2017self}     & 0.224 & 0.129 & 0.089 & 0.068 & 0.308 & 0.297\\
 & CoAtt~\cite{jing2017automatic}   & 0.455 & 0.288 & 0.205 & 0.154 & 0.369 & 0.277\\
 & HGRG-Agent~\cite{li2018hybrid}   & 0.438 & 0.298 & 0.208 & 0.151 & 0.322 & \textbf{0.343}\\
 \cline{2-8}
 & CMAS$_\text{W}$     & 0.440 & 0.292 & 0.204 & 0.147 & 0.365 & 0.252\\
 & CMAS$_\text{NW,AW}$  & 0.451 & 0.286 & 0.199 & 0.146 & 0.366 & 0.269\\
 & CMAS-IL       & 0.454 & 0.283 & 0.195 & 0.143 & 0.353 & 0.266\\
 & CMAS-RL       & \textbf{0.464} & \textbf{0.301} & \textbf{0.210} & \textbf{0.154} & \textbf{0.362} & 0.275\\
\hline
\end{tabular}
\caption{Main results for findings generation on the CX-CHR (upper) and IU-Xray (lower) datasets. BLEU-n denotes the BLEU score that uses up to n-grams. }
\label{tab:main_results}
\end{table*}

\begin{table*}[t!]
\centering
\small
\begin{tabular}{c|l|c c c c c c}
\hline
Dataset & Methods & BLEU-1 & BLEU-2 & BLEU-3 & BLEU-4 & ROUGE & CIDEr\\
\hline
\multirow{6}{*}{CX-CHR}
 & SoftAtt$_\text{text}$~\cite{xu2015show}    & 0.112 & 0.044 & 0.016 & 0.005 & 0.142 & 0.038\\
 & SoftAtt$_\text{vision}$~\cite{xu2015show}   & 0.408 & 0.300 & 0.247 & 0.208 & 0.466 & 0.932\\
 \cline{2-8}
 & CMAS$_\text{text}$     & 0.182 & 0.141 & 0.127 & 0.119 & 0.356 & 2.162\\
 & CMAS$_\text{vision}$  & 0.415 & 0.357 & \textbf{0.323} & \textbf{0.296} & \textbf{0.511} & \textbf{3.124}\\
 & CMAS-IL       & 0.426 & 0.360 & 0.322 & 0.290 & 0.504 & 3.080\\
 & CMAS-RL       & \textbf{0.428} & \textbf{0.361} & \textbf{0.323} & 0.290 & 0.504 & 2.968\\
\hline
\hline
\multirow{6}{*}{IU-Xray}
 & SoftAtt$_\text{text}$~\cite{xu2015show}    & 0.179 & 0.047 & 0.006 & 0.000 & 0.161 & 0.032\\
 & SoftAtt$_\text{vision}$~\cite{xu2015show}   & 0.224 & 0.103 & 0.045 & 0.022 & 0.210 & 0.046\\
 \cline{2-8}
 & CMAS$_\text{text}$     & 0.316 & 0.235 & 0.187 & 0.148 & \textbf{0.537} & \textbf{1.562}\\
 & CMAS$_\text{vision}$  & 0.379 & 0.270 & 0.203 & 0.151 & 0.513 & 1.401\\
 & CMAS-IL       & 0.399 & 0.285 & 0.214 & 0.158 & 0.517 & 1.407\\
 & CMAS-RL       & \textbf{0.401} & \textbf{0.290} & \textbf{0.220} & \textbf{0.166} & 0.521 & 1.457\\
\hline
\end{tabular}
\caption{Main results for impression generation on the CX-CHR (upper) and IU-Xray (lower) datasets. BLEU-n denotes the BLEU score that uses up to n-grams.}
\label{tab:main_results2}
\end{table*}

\subsection{Main Results}\label{main_results} 

\paragraph{Comparison to State-of-the-art}
Table~\ref{tab:main_results} shows results on the automatic metrics for the Findings module.
On both datasets, CMAS outperforms all baseline methods on almost all metrics, which indicates its overall efficacy for generating reports that resemble those written by human experts. 
The methods can be divided into two different groups: single sentence models (CNN-RNN, LRCN, AdaAtt, Att2in) and hierarchical models (CoAtt, HGRG-Agent, CMAS).
Hierarchical models consistently outperform single sentence models on both datasets, suggesting that the hierarchical models are better for modeling paragraphs. 
The leading performances of CMAS-IL and CMAS-RL over the rest of hierarchical models demonstrate the validity of our practice in exploiting the structure information within sections.

\begin{table*}[h!]
    \centering
    \scriptsize
    \begin{tabular}{c|c|c|c|c|c|c|c|c}
    \hline
    Dataset & \multicolumn{4}{c}{CX-CHR} & \multicolumn{4}{|c}{IU-Xray}\\
    \hline
    Methods & \newcite{li2018hybrid} & CMAS$_\text{NW,AW}$ & CMAS-IL & CMAS-RL & \newcite{li2018hybrid} & CMAS$_\text{NW,AW}$ & CMAS-IL & CMAS-RL \\
    \hline
    Precision  & 0.292 & 0.173 & 0.272 & \textbf{0.309} & 0.121 & 0.070 & 0.094 & \textbf{0.128}\\
    \hline
    FPR  & 0.059 & 0.076 & 0.063 & \textbf{0.051} & 0.043 & 0.044 & 0.012 & \textbf{0.007}\\
    \hline
    \end{tabular}
    \caption{Average precision and average False Positive Rate (FPR) for abnormality detection. (\emph{Findings})}
    \label{tab:acc}
\end{table*}

\begin{table*}[h!]
    \centering
    \scriptsize
    \begin{tabular}{c|c|c|c|c|c|c|c|c}
    \hline
    Dataset & \multicolumn{4}{c}{CX-CHR} & \multicolumn{4}{|c}{IU-Xray}\\
    \hline
    Methods & CMAS$_\text{text}$ & CMAS$_\text{vision}$ & CMAS-IL & CMAS-RL & CMAS$_\text{text}$ & CMAS$_\text{vision}$ & CMAS-IL & CMAS-RL \\
    \hline
    Precision  & 0.067 & 0.171 & 0.184 & \textbf{0.187} & 0.054 & 0.160 & 0.162 & \textbf{0.165}\\
    \hline
    FPR  & \textbf{0.067} & 0.142 & 0.170 & 0.168 & \textbf{0.023} & 0.024 & 0.024 & 0.024\\
    \hline
    \end{tabular}
    \caption{Average precision and average False Positive Rate (FPR) for abnormality detection. (\emph{Impression})}
    \label{tab:acc2}
\end{table*}

\paragraph{Ablation Study}
$\text{CMAS}_\text{W}$ has only one writer, which is trained on both normal and abnormal findings. 
Table~\ref{tab:main_results} shows that $\text{CMAS}_\text{W}$ can achieve competitive performances to the state-of-the-art methods. 
$\text{CMAS}_\text{NW, AW}$ is a simple concatenation of two single agent models $\text{CMAS}_\text{NW}$ and $\text{CMAS}_\text{AW}$, where $\text{CMAS}_\text{NW}$ is trained only on normal findings and $\text{CMAS}_\text{AW}$ is trained only on abnormal findings.
At test time, the final paragraph of $\text{CMAS}_\text{NW, AW}$ is simply a concatenation of normal and abnormal findings generated by $\text{CMAS}_\text{NW}$  and $\text{CMAS}_\text{AW}$ respectively.
Surprisingly, $\text{CMAS}_\text{NW, AW}$ performs worse than $\text{CMAS}_\text{W}$ on the CX-CHR dataset.
We believe the main reason is the missing communication protocol between the two agents, which could cause conflicts when they take actions independently.
For example, for an image, NW might think ``the heart size is normal'', while AW believes ``the heart is enlarged''.
Such conflict would negatively affect their joint performances. 
As evidently shown in Table~\ref{tab:main_results}, CMAS-IL achieves higher scores than $\text{CMAS}_\text{NW, AW}$, directly proving the importance of communication between agents and thus the importance of PL.
Finally, it can be observed from Table~\ref{tab:main_results} that CMAS-RL consistently outperforms CMAS-IL on all metrics, which demonstrates the effectiveness of reinforcement learning.

\paragraph{Impression Module}
As shown in Table~\ref{tab:main_results2}, CMAS$_\text{vision}$ and CMAS$_\text{text}$ have higher scores than SoftAtt$_\text{vision}$ and SoftAtt$_\text{text}$, indicating the effectiveness of CMAS.
It can also be observed from Table~\ref{tab:main_results2} that images provide better information than text, since CMAS$_\text{vision}$ and SoftAtt$_\text{vision}$ exceed the scores of CMAS$_\text{text}$ and SoftAtt$_\text{text}$ to a large margin on most of the metrics.
However, further comparison among CMAS-IL, CMAS$_\text{text}$ and CMAS$_\text{vision}$ shows that text information can help improve the model's performance to some degree.

\begin{figure*}[h!]
    \centering
    \includegraphics[width=1.\textwidth]{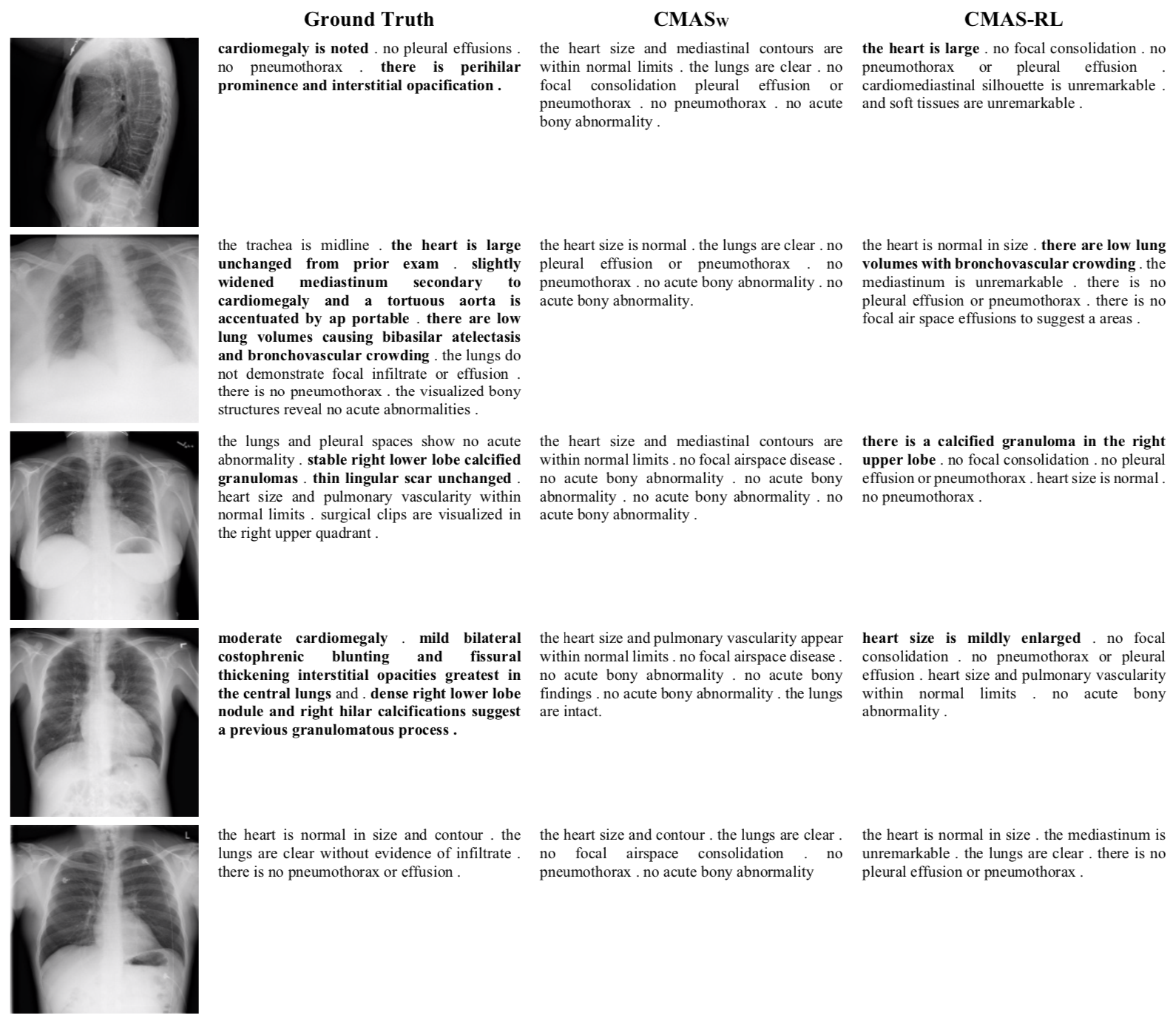}
    \caption{Examples of findings generated by CMAS-RL and CMAS$_\text{W}$ on IU-Xray dataset, along with their corresponding CXR images and ground-truth reports. 
    Highlighted sentences are abnormal findings.}
    \label{fig:samples}
\end{figure*}

\subsection{Abnormality Detection}
The automatic evaluation metrics (e.g. BLEU) are based on n-gram similarity between the generated sentences and the ground-truth sentences.
A model can easily obtain high scores on these automatic evaluation metrics by generating normal findings~\cite{jing2017automatic}.
To better understand CMAS's ability in detecting abnormalities, we report its precision and average False Positive Rate (FPR) for abnormality term detection in Table~\ref{tab:acc} and Table~\ref{tab:acc2}. 
Table~\ref{tab:acc} shows that CMAS-RL obtains the highest precision and the lowest average FPR on both datasets, indicating the advantage of CMAS-RL for detecting abnormalities.
Table~\ref{tab:acc2} shows that CMAS-RL achieves the highest precision scores, but not the lowest FPR. 
However, FPR can be lowered by simply generating normal sentences, which is exactly the behavior of CMAS$_\text{text}$. 

\subsection{Qualitative Analysis}
In this section, we evaluate the overall quality of generated reports through several examples.
Figure~\ref{fig:samples} presents 5 reports generated by CMAS-RL and CMAS$_\text{W}$, where the top 4 images contain abnormalities and the bottom image is a normal case.
It can be observed from the top 4 examples that the reports generated by CMAS-RL successfully detect the major abnormalities, such as ``cardiomegaly'', ``low lung volumes'' and ``calcified granulomas''. 
However, CMAS-RL might miss secondary abnormalities sometimes. 
For instance, in the third example, the ``right lower lobe'' is wrongly-written as ``right upper lobe'' by CMAS-RL. 
We find that both CMAS-RL and $\text{CMAS}_\text{W}$ are capable of producing accurate normal findings since the generated reports highly resemble those written by radiologists (as shown in the last example in Figure~\ref{fig:samples}). 
Additionally,  $\text{CMAS}_\text{W}$ tends to produce normal findings, which results from the overwhelming normal findings in the dataset.

\subsection{Template Learning}
Radiologists tend to use reference templates when writing reports, especially for normal findings. 
Manually designing a template database can be costly and time-consuming.
By comparing the most frequently generated sentences by CMAS with the most used template sentences in the ground-truth reports, we show that the Normality Writer (NW) in the proposed CMAS is capable of learning these templates automatically.
Several most frequently used template sentences~\cite{li2018hybrid} in the IU-Xray dataset are shown in Table~\ref{tab:gt_template}.
The top 10 template sentences generated by NW are presented in Table~\ref{tab:learned_template}. 
In general, the templates sentences generated by NW are similar to those top templates in ground-truth reports.

\begin{table}[h!]
    \centering
    \scriptsize
    \begin{tabular}{l}
    \hline
     The lungs are clear.   \\
     Lungs are clear. \\
     The lung are clear bilaterally.\\
    \hline
    No pneumothorax or pleural effusion. \\
    No pleural effusion or pneumothorax. \\
    There is no pleural effusion or pneumothorax.\\
    \hline
    No evidence of focal consolidation, pneumothorax, or pleural effusion.\\
    No focal consolidation, pneumothorax or large pleural effusion. \\
    No focal consolidation, pleural effusion, or pneumothorax identified..\\
    \hline
    \end{tabular}
    \caption{Most commonly used templates in IU-Xray. 
    Template sentences are clustered by their topics.}
    \label{tab:gt_template}
\end{table}

\begin{table}[h!]
    \centering
    \scriptsize
    \begin{tabular}{l}
    \hline
    The lungs are clear. \\
    \hline
    The heart is normal in size. \\
    Heart size is normal.\\
    \hline
    There is no acute bony abnormality. \\
    \hline
    There is no pleural effusion or pneumothorax. \\
    There is no pneumothorax.\\
    No pleural effusion or pneumothorax.\\
    \hline
    There is no focal air space effusion to suggest a areas.\\
    No focal consolidation.\\
    Trachea no evidence of focal consolidation pneumothorax or pneumothorax.\\
    \hline
    \end{tabular}
    \caption{Top 10 sentences generated by CMAS. 
    The sentences are clustered by their topics.}
    \label{tab:learned_template}
\end{table}

\section{Conclusion}\label{conclusion}
In this paper, we proposed a novel framework for accurately generating chest X-ray imaging reports by exploiting the structure information in the reports.
We explicitly modeled the between-section structure by a two-stage framework, and implicitly captured the within-section structure with a novel Co-operative Multi-Agent System (CMAS) comprising three agents: Planner (PL), Abnormality Writer (AW) and Normality Writer (NW).
The entire system was trained with REINFORCE algorithm.
Extensive quantitative and qualitative experiments demonstrated that the proposed CMAS not only could generate meaningful and fluent reports, but also could accurately describe the detected abnormalities.






\bibliography{acl2019}
\bibliographystyle{acl_natbib}
\end{document}